# APPLICATION OF ARTIFICIAL NEURAL NETWORKS IN ESTIMATING PARTICIPATION IN ELECTIONS

Seyyed Reza Khaze[1], Mohammad Masdari[2] and Sohrab Hojjatkhah[3]

[1]Department of Computer Engineering, Dehdasht Branch, Islamic Azad University, Iran,
khaze@iaudehdasht.ac.ir, khaze.reza@gmail.com

[2]Department of Computer Engineering, Urmia Branch, Islamic Azad University, Iran,
m.masdari@iaurmia.ac.ir

[3]Department of Computer Engineering, Dehdasht Branch, Islamic Azad University, Iran,
hojjatkhah@gmail.com

## ABSTRACT

*It is approved that artificial neural networks can be considerable effective in anticipating and analyzing flows in which traditional methods and statics aren't able to solve. In this article, by using two-layer feed-forward network with tan-sigmoid transmission function in input and output layers, we can anticipate participation rate of public in kohgiloye and Boyerahmad Province in future presidential election of Islamic Republic of Iran with 91% accuracy. The assessment standards of participation such as confusion matrix and ROC diagrams have been approved our claims.*

## 1. INTRODUCTION

Anticipating the political behavior of the public will demonstrate the future landscape of a society in which everyone can experience the future space virtually and begin to program and make policies about his/her job career, political, economic, and military and other sections affairs being informed about available characteristics, parties and candidates viewpoints. As many political theorists have a lot of problems in anticipating public political behavior in these kinds of elections and aren't able to do it accurately using statistical tools and social and political analyzing methods [1], so, data mining methods such as artificial neural networks are used to find potential rules in these kinds of data by gathering information and necessary documents of people previous electoral and political behaviors [2].

Artificial neural network is considered as one of the modern mathematical-computational methods which are used to solve un-anticipated dynamic problems in developed behavioral systems during a time period. By learning to recognize patterns from data in which other computational and statistical method failed to solve them, artificial neural networks are able to solve the problems. Nowadays, artificial neural networks are widely used due to their ability to anticipate. Considering that this feature , participation anticipate and election results are among the most interesting challenges for political scientists, it can be used artificial neural network to anticipate in political issues, although it was less used in Political Science by scholars. Political scientists and scholars are already search for vast models to anticipate election results throughout developing world.

They use survey models which are a series of indicators to construct analytical model of electoral behavior, local election analyses and previous election results survey which isn't possible to anticipate accurately due to possible nature problems and uncertainty of new electoral behaviors and results [3]. Artificial neural networks can be seen as an approximation of the actual neural network because it simulates how biological neurons act in the human brain structure in which the same processing units of neurons are connected to each other by synapses. These networks are a set of interconnected processing units that analyze the model in which a set of input data mapped out into a set of output units. The artificial neural networks are used to anticipate and classify [4].

In the first part of this paper reviews the applications of artificial neural networks to predict. In the next section, the Estimating Participation in Elections results using these networks have been discussed and then participate in the election of the case study is discussed. This is done using a two-layer feed-forward network. And finally discuss the evaluation of the performance of artificial neural networks.

## 2. PREDICT USING NEURAL NETWORKS

The increasing tendency to anticipate using artificial neural networks resulted in a remarkable increase in research activities in the recent decade [5]. Artificial neural networks are suitable methods to indicate their efficiency in anticipating exchange rate, analyzing economical time series, issues related to stock and stock market. Results and the function of artificial neural networks have shown that this model has better function than popular methods considering assessment criteria of future time sequences anticipation. In recent years, many complicated statistical methods have been developed and used for anticipation process in the related issues. However, there are two basic problems about these methods. It includes personal statistical problem, power and certain analyses for a single and multi-dimensional time series. Artificial neural networks were reliable methods to remove statistical problems in anticipating multi-dimensional time series [6].

In the economic field, it is hard to anticipate macro-economic issues due to the lack of accurate and satisfying model. The most accurate model for economical anticipating is the time series model of black-box model which partly considers the economic structure. High noise levels, short time series, and nonlinear effects are introduced as time series problems solutions of regression method which are the popular ways to solve economical anticipations in the recent years. The given solution of artificial neural networks can be used to solve noted problems to anticipate macro-economic issues. By selecting the advanced parameters in artificial networks, the researchers have been provided the experimental results to anticipate the indicator of industrial production in USA in which the given results of their simulation experiments indicate the better function of artificial neural networks than popular linear time series and regression [7].

In political field, the flexible model of neural networks is capable of recognizing and anticipating the empirical relationship between democracy and global collision theory while the logical model is unable to do so. The logic model was the proposed model of Marchi et al. They claimed it is superior than artificial neural network which rejected by Nathaniel beck et al. They began to basically analyze Grynaviski, Gelpi & Marchi theory and emphasized that their theories are completely in common with theirs about the world. Their common viewpoint is using main standards to assess the world studies in order to anticipate out of sample functions. They also indicated that other conclusions of Marchi et al are incorrect. They approved that using same assessment criterion for each two models which Marchi et al had been used was based on logic model superiority to the other neural networks which provided challenges and rejected their theory [8].

In tourism field, artificial neural network model had better function to anticipate international tourism demands than other popular anticipation methods, time series and regression techniques. It is approved in a research executed by Rob Law & Norman Au. The authors have been used organized Feed-Forward network which is a particular model of neural network to anticipate Japanese trip to Hong-Kong. Networks consisting of input and output layers in which input layer includes 6 service price nodes, average hotel rates, exchange rates, population, internal gross expenditures and output layer of a node is to represent a travel demands from Japan to Hong - Kong. The achieved experimental results obtained from the study of the time interval from 1967 to 1996 has shown that the neural network model has better results than other methods [9].

In Environmental area, neural network can be used to solve available problems and issues of processes modeling of real world by designing composed architecture with meta-heuristic algorithms. These problems include nonlinear multidimensional space and chaos theory. Although, designing neural networks for high efficiency is difficult, but Harry Niska et al have been used multi-layer perceptron model to anticipate Nitrogen Dioxide during day time in a crowded traffic station in Helsinki city by parallel combining of genetic algorithm in parallel to choose input and design high level architecture. It removed the practical problems of designing neural networks for high efficiency and indicated that combined networks can guarantee it [10].

In informational technology field, neural networks have also shown that short-term intervals can provide better function than adapted methods of time series to anticipate traffic rate in top-in networks. Paulo Cortez et al have been executed several experiments to assess anticipation accuracy in the noted methods using real-world data and related information to internet service providers. The experiments have been done in separated time sequences of 5 min, 1 h and a day. However, time series of 1-day anticipation had better results but the neural group experiments have shown that it can be provided best results for short-term time series [11].

## 3. ESTIMATING PARTICIPATION IN ELECTIONS USING NEURAL NETWORK

Multilayer Feed-Forward is an example of artificial neural networks in which learning of a neural network is performed by back propagation algorithm. It frequently repeats the weight learning process to classify and anticipate of class from tuples. A Multilayer Feed-Forward network consists of an input layer, one or more hidden layer and output layer. Each layer consists of units. The input network is in the size of features which have educational samples and enters to the input layer, simultaneously. These inputs are transported through input layer and then become weighted. Then, they are transported to the second layer of pseudo-neuron units which known as hidden layer. The output of hidden layer can be entered to another hidden layer as an input of output layer. The weighted output or hidden layers are used as input of output layer in which they can reach to the ideal weight and network learning by educational data series and provide classification and anticipation operation for tuples and test samples [12].We suppose to anticipate participation rate in 11th election of Islamic Republic of Iran in Kohkiloye and Boyerahmad Province using artificial neural network. These data includes features and prospects of 100 qualified persons to attend in election [1] which is shown its fields in Tables 1-9.

Table 1.Age of people

| ID | Label |
|----|-------|
| 1 | Old |

| 2 | Middle-aged |
| 3 | Young |

Table 2.Degree of people

| ID | Label |
|---|---|
| 1 | PhD |
| 2 | Masters |
| 3 | BA |
| 4 | Under license |

Table 3.Jobs of People

| ID | Label |
|---|---|
| 1 | Government employees |
| 2 | Professors and university staff |
| 3 | Free jobs |

Table 4.Political Orientation of people

| ID | Label |
|---|---|
| 1 | Reformists |
| 2 | Fundamentalists |
| 3 | "Velayi" |

Table 5.Opinion of people about government services

| ID | Label |
|---|---|
| 1 | Mortgage |
| 2 | targeted subsidies |
| 3 | Marriage Loans |
| 4 | Fuel Rationing |

Table 6: Opinion of people about Participation type in elections

| ID | Label |
|---|---|
| 1 | Religious duty and religious |
| 2 | Participation in decision making |
| 3 | Overall reform |

Table 7.Opinion of people about general policy in international affairs

| ID | Label |
|---|---|
| 1 | Resistance |
| 2 | Negotiation |
| 3 | Concessions |

Table 8.Opinion of people about the election officials

| ID | |
|---|---|
| 1 | Higher accuracy |
| 2 | Confidence |
| 3 | Unreliability |

Table 9.Opinion of people about Candidates

| ID | Label |
|---|---|
| 1 | Popular candidate |
| 2 | Party candidate |
| 3 | Political elite |

Table 10.Opinion of people about Participation in elections

| ID | |
|---|---|
| 1 | Partnership |
| 2 | possible participation |
| 3 | Without participation |

We, firstly, create an artificial neural network using MATLAB tools. Our artificial neural network is a Feed-Forward network with tan-sigmoid transmission function in the hidden and output layers. In this network, we use 10 neurons in hidden layer. The network has 10 inputs and 3 outputs as the target vector has 3 members.

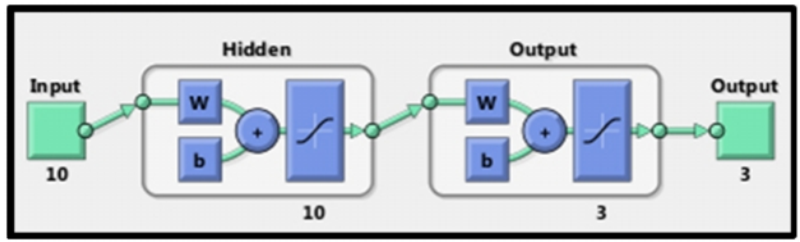

Figure 1.Overall architecture of the Feed-Forward Neural Network to Predict Participation

We provide the number of hidden layers equal to 10 and create it based on 10 hidden layers. In pre-process stage, the data of similar additional rows are removed if it is available. In the third stage, we classify all input data to groups, randomly and use some data for educational sample, assessment and the remains for network test. The number and percentage of each stage are shown in Table 11.

Table 11.number and percentage of each stage

| Stage | Sample |
|---|---|
| Training | 70 |
| Validation | 15 |
| Testing | 15 |

# 4. DISCUSSION

In network learning part, we begin to learn it by choosing Gradient learning function. It is necessary to note that Mean square error function is selected to recognize network efficiency. To find performed anticipation rate and real values, it is used error function concept. It is also used mean square error to assess anticipation error [13, 14]. If the assessment error repeated 6 consecutive times, the learning procedure must be stopped. In this case, it is occurred in 15 repetitions. The diagram of learning errors, assessment errors and test errors are shown in Fig

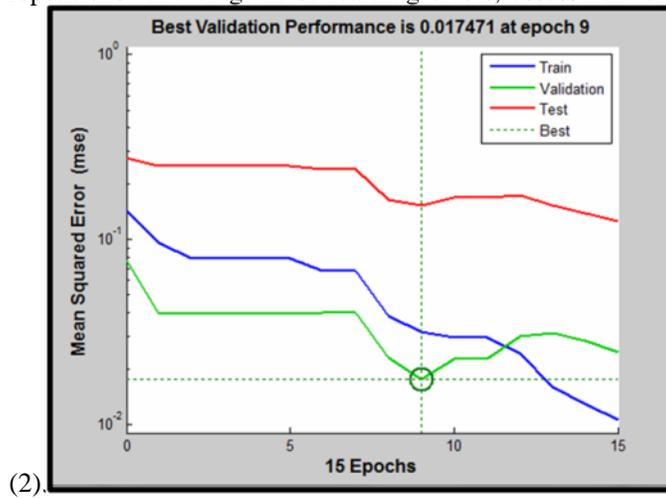

(2).

Figure 2.Diagram of training, evaluation and testing error of Feed-Forward Neural Network to Predict Participation

It is clear that the best efficiency is occurred in repetition 9. The learning function of learning data is shown in repetition 15 in Fig 3

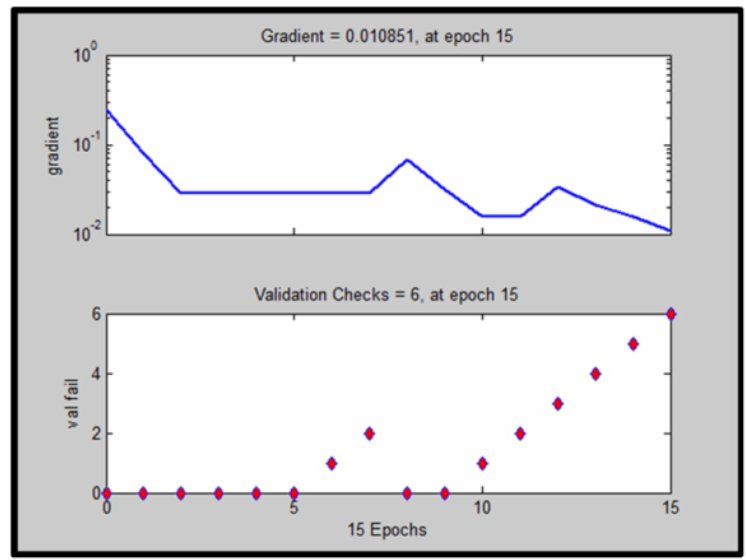

Figure 3.Training Performance of Feed forward Neural Networks to Predict Participation

In Fig 4, the histogram of learning stage errors are indicated.

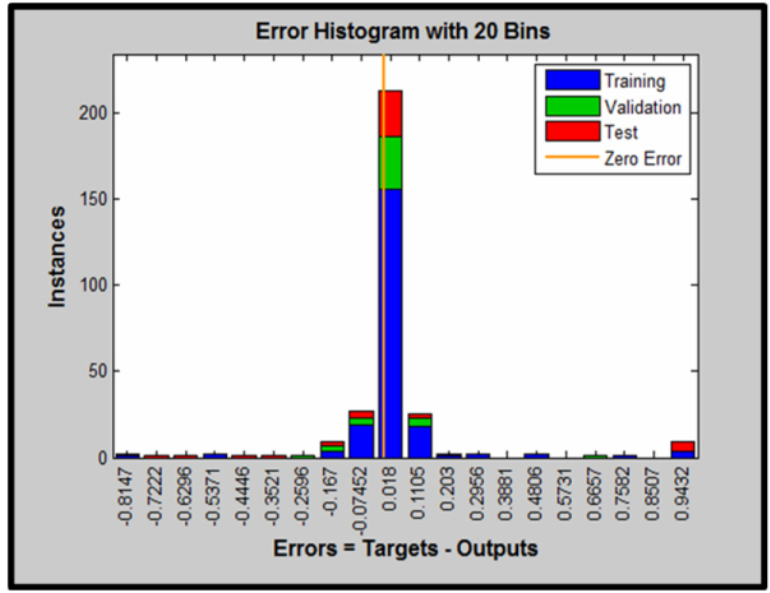

Figure 4.Histogram of errors in training

In assessments, it is also used confusion matrix which indicates the classified samples of a class in another class or similar one [15]. The reaction of distribution matrix analyses and different errors about the final network are indicated in Fig 5

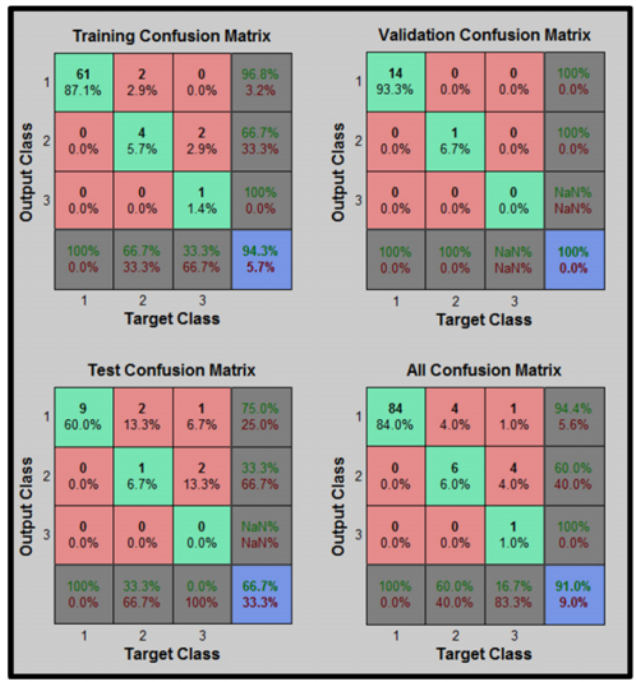

Figure 5.Collision Matrix to predict participation

As it can be seen, the used neural network to anticipate participation result in election can provide 91% accuracy in future election of Islamic Republic of Iran in Kohkiloye and Boyerahmad Province. The ROC curve also indicates the real positive rate and false positive rate. False positive rate is a proportion of negative tuples which recognize positive wrongly and provide for a model [13, 14]. Diagonal elements in each table indicate the number of cases which classified correctly. The non-diagonal elements are those which classified wrongly. The blue-colored cell (in the lower right corner of the table) indicates the total wrong classification percent rate   in red color. In ROC curve, the positive accurate percent rate is outlined opposite of wrong positive percent rate

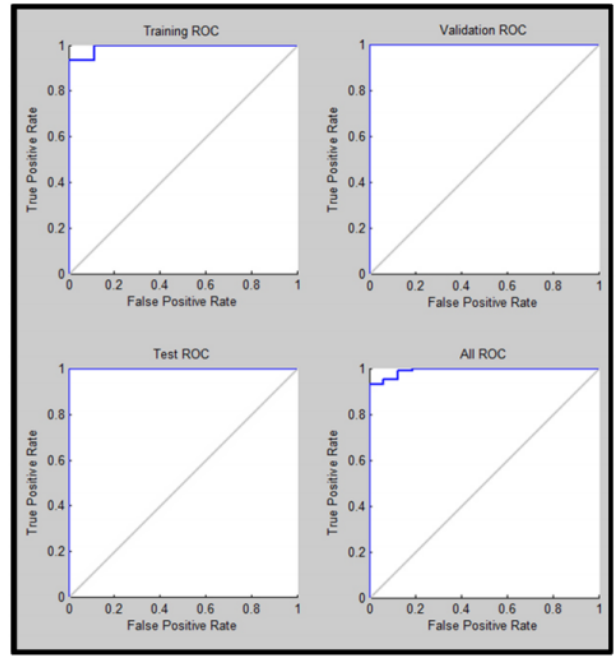

Figure 6.ROC curve for prediction of participation

As it can be seen in ROC diagrams, neural network has good classification function.

## 5. CONCLUSIONS

Artificial neural network is considered as one of the modern mathematical-computational methods which are used to solve un-anticipated dynamic problems in developed behavioral systems during a time period. Nowadays, artificial neural networks are widely used due to their ability to anticipate. Considering that this feature , participation anticipate and election results are among the most interesting challenges for political scientists, it can be used artificial neural network to anticipate in political issues. We use artificial neural networks to predict participation in presidential elections in Iran. Our data consist of 100 individual attitudes of Dehdasht citizenship. We use a neural network feed forward neural network with tan-sigmoid transfer function. We have been able to participate in the election with 91 percent accuracy to predict. ROC graphs and other standards of our claim are proved.

## 9. REFERENCES

[1]   Sangar, A. B., Khaze, S. R., & Ebrahimi, L (2013), Participation Anticipating In Elections Using Data Mining Methods , International Journal On Cybernetics & Informatics,2(2),47-60.

[2] Gharehchopogh, F. S. & Khaze, S. R. (2012). Data Mining Application for Cyber Space Users Tendency In Blog Writing: A Case Study, International journal of computer applications, 47(18), 40-46.

[3] Gill, G. S. (2008), Election Result Forecasting Using Two Layer Perceptron Network ,Journal of Theoretical and Applied Information Technology, 47(11),1019-1024

[4] Caleiro, A. (2005), How to Classify a Government? Can a Neural Network do it? , University of Evora, Economics Working Papers.

[5] Adya, M., & Collopy, F. (1998), How Effective Are Neural Networks at Forecasting and Prediction? A Review and Evaluation, J. Forecasting, 17, 481-495.

[6] Thiesing, F.M., Vornberger, O. (1997), Sales Forecasting Using Neural Networks, Neural Networks, International Conference On ,4, 2125-2128 .

[7] Moody, J. (1995), Economic Forecasting: Challenges and Neural Network Solutions, In Proceedings of the International Symposium on Artificial Neural Networks

[8] Beck, N., King, G., & Zeng, L. (2004), Theory and Evidence in International Conflict: A Response to De Marchi, Gelpi, and Grynaviski, American Political Science Review, 98(2), 379-389.

[9] Law, R., & Au, N. (1999), A Neural Network Model To Forecast Japanese Demand For Travel To Hong Kong, Tourism Management, 20(1), 89-97.

[10] Niska, H., Hiltunen, T., Karppinen, A., Ruuskanen, J., & Kolehmainen, M. (2004), Evolving The Neural Network Model For Forecasting Air Pollution Time Series, Engineering Applications of Artificial Intelligence, 17(2), 159-167.

[11] Cortez, P., Rio, M., Rocha, M., & Sousa, P. (2012), Multi-Scale Internet Traffic Forecasting Using Neural Networks And Time Series Methods, Expert Systems, 29(2), 143-155.

[12] Towell, G. G., & Shavlik, J. W. (1994), Knowledge-Based Artificial Neural Networks, Artificial intelligence, 70(1), 119-165.

[13] Gorunescu, F. (2011), Data Mining: Concepts, Models and Techniques (Vol. 12), Springerverlag Berlin Heidelberg.

[14] Han, J., Kamber, M., & Pei, J. (2006), Data Mining: Concepts And Techniques. Morgan Kaufmann.

[15] Yang, S., Yan, J., Gao, C., & Tan, G. (2011), Blogger's Interest Mining Based On Chinese Text Classification, In Nonlinear Mathematics for Uncertainty and its Applications (pp. 611-618), Springer Berlin Heidelberg.

## Authors

**Seyyed Reza Khaze** is a Lecturer and Member of the Research Committee of the Department of Computer Engineering, Dehdasht Branch, Islamic Azad University, Iran. He is a Member of Editorial Board and Review Board in Several International Journals and National Conferences. His interested research areas are in the Software Cost Estimation, Machine learning, Data Mining, Optimization and Artificial Intelligence.

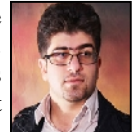

**Mohammad Masdari** is Currently Head of the Department of Computer Engineering, Urmia Branch, Islamic Azad University, Iran. He Has a Currently PhD Candidate In Department Of Computer Engineering At Science And Research Branch, Islamic Azad University, Iran. His Interested Research Areas Are in the Wireless Sensor Networks, data mining, optimization and artificial intelligence.

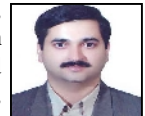

**Sohrab Hojjatkhah** is Currently Head of the Department of Computer Engineering, Dehdasht Branch, Islamic Azad University, Iran. He Has a Bachelor's Degree In Software Engineering, Received From Amir Kabir University, Iran, Then Received A Master's Degree In Artificial Intelligence From The University Of Shiraz And Currently PhD Candidate In Department Of Computer Engineering At Science And Research Branch, Islamic Azad University, Iran. His Interested Research Areas Are In The Image Processing, Speech Processing, Machine Learning, Data Mining And Artificial Intelligence.

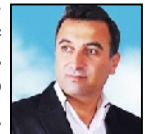